\documentclass{article}

\PassOptionsToPackage{numbers, compress}{natbib}
\usepackage[preprint]{neurips_2026}

\usepackage[utf8]{inputenc} % allow utf-8 input
\usepackage[T1]{fontenc}    % use 8-bit T1 fonts
\usepackage{hyperref}       % hyperlinks
\usepackage{url}            % simple URL typesetting
\usepackage{booktabs}       % professional-quality tables
\usepackage{amsfonts}       % blackboard math symbols
\usepackage{nicefrac}       % compact symbols for 1/2, etc.
\usepackage{microtype}      % microtypography
\usepackage{graphicx}
\usepackage{amsmath}
\usepackage{amssymb}
\usepackage{array}
\usepackage{enumitem}
\usepackage{algorithm}
\usepackage[noend]{algpseudocode}
\usepackage{multirow}
\usepackage{listings}
\usepackage{xcolor}
\usepackage{caption}
\usepackage{wrapfig}
\definecolor{listaccent}{HTML}{3E7C59}   % ACCENT (frame)
\definecolor{listaccentbg}{HTML}{EAF3EE} % ACCENT_BG (background)
\definecolor{listink}{HTML}{262626}      % INK (keys/values)
\definecolor{listnum}{HTML}{B0332A}      % GAP (numbers/booleans/null)
\definecolor{listpunct}{HTML}{8C8C8C}    % muted punctuation
\definecolor{diffadd}{HTML}{1B6E3C}      % added lines
\definecolor{diffdel}{HTML}{A32B22}      % removed lines
\definecolor{diffhunk}{HTML}{5B4E8C}     % hunk headers
\lstdefinelanguage{skilldiff}{
  basicstyle=\ttfamily\scriptsize\color{listink},
  breaklines=true,
  breakindent=0pt,
  columns=fullflexible,
  showstringspaces=false,
  frame=single,
  rulecolor=\color{listaccent},
  framerule=0.8pt,
  backgroundcolor=\color{listaccentbg},
  framesep=5pt,
  xleftmargin=3pt,
  xrightmargin=3pt,
  morecomment=[f][\color{diffdel}]{-},
  morecomment=[f][\color{diffadd}]{+},
  morecomment=[f][\color{diffhunk}\itshape]{@@},
}
\lstdefinelanguage{json}{
  basicstyle=\ttfamily\scriptsize\color{listink},
  breaklines=true,
  columns=fullflexible,
  showstringspaces=false,
  frame=single,
  rulecolor=\color{listaccent},
  framerule=0.8pt,
  backgroundcolor=\color{listaccentbg},
  framesep=5pt,
  xleftmargin=3pt,
  xrightmargin=3pt,
  stringstyle=\color{listink},
  commentstyle=\color{listpunct},
  keywordstyle=\bfseries\color{listnum},
  morestring=[b]",
  literate=
    *{0}{{{\color{listnum}0}}}{1}
     {1}{{{\color{listnum}1}}}{1}
     {2}{{{\color{listnum}2}}}{1}
     {3}{{{\color{listnum}3}}}{1}
     {4}{{{\color{listnum}4}}}{1}
     {5}{{{\color{listnum}5}}}{1}
     {6}{{{\color{listnum}6}}}{1}
     {7}{{{\color{listnum}7}}}{1}
     {8}{{{\color{listnum}8}}}{1}
     {9}{{{\color{listnum}9}}}{1},
}

\title{Skill Transfer: Adapting Agent Skills from Strong to Weak Agents}

\author{
Yifan Lan,
Hanyu Wang,
Lu Lin,
Jinghui Chen \\
The Pennsylvania State University \\
\texttt{\{yifanlan,hbw5365,lxl5598,jzc5917\}@psu.edu}
}

\newif\ifcomments
\commentstrue

\setlist{topsep=3pt, itemsep=1pt, parsep=0pt, partopsep=0pt}

\begin{document}

\maketitle

\begin{abstract}
    Agent skills are modular bundles of instructions, code, and resources that an agent loads on demand, increasingly authored and validated on frontier models and shared as an open standard. The same skill behaves differently across models: a strong model turns it into a large performance gain, while a weak model realizes little of that gain and sometimes does worse than with no skill at all. Cost and resource constraints, however, often restrict real deployments to exactly these weaker agents. We hypothesize that the strong model silently supplies procedural steps the skill leaves implicit, which the weak model omits, and formalize closing this gap as \emph{Skill Transfer}: holding the skill's capability ceiling and the model's parameters fixed, we adapt the skill so a specific weak agent can execute it well. Our method abstracts strong and weak trajectories into typed execution structures, localizes their structural divergence with a graph edit distance (GED), and repairs the skill through an iterative diagnoser-patcher loop. Across three benchmarks and two model families, Skill Transfer raises pass rate over the unmodified skill significantly and well over other skill-evolution baselines. The adapted skill also improves the strong agent it was never adapted for, and transfers out of distribution to a structurally related, unseen task. Code is available at \url{https://github.com/Yifan-Lan/Skill-Transfer}.
\end{abstract}

\section{Introduction}

Large Language Model agents solve complex tasks by interleaving reasoning~\citep{cot,selfconsistency,tot,illusion-reasoning} with actions over an external environment~\citep{react,toolformer,sweagent}, demonstrating excellent performance. However, they often lack knowledge of some specific domains where fine-tuning is expensive and harmful to general capability. 
% \why{where fine-tuning is expensive...}. 
% A growing practice to solve this packages the workflow and knowledge for a domain into a reusable skill: a self-contained bundle of natural-language instructions, runnable scripts, and reference material that the agent loads only when a task matches~\citep{anthropic-skills} 
To solve this, agent skills have been proposed, serving as reusable file systems where procedural domain knowledge is structurally packaged~\citep{anthropic-skills}. Specifically, an agent skill is a self-contained bundle of natural-language instructions, runnable scripts, and reference material that the agent loads only when a task matches. Skills are compositional, interpretable, and portable. An agent can accumulate a library of them and reuse it on unseen tasks~\citep{voyager,skillrl,trace2skill}, and recent open standards let a skill authored once be shared across different agents and providers~\citep{anthropic_skills,openai_skills}. 
% The shared skill is expected to work effectively on all agents.
By design, skills are expected to work effectively across various agents, agnostic to specific harnesses and models. 

\begin{figure*}[t]
    \centering
    \includegraphics[width=1\linewidth]{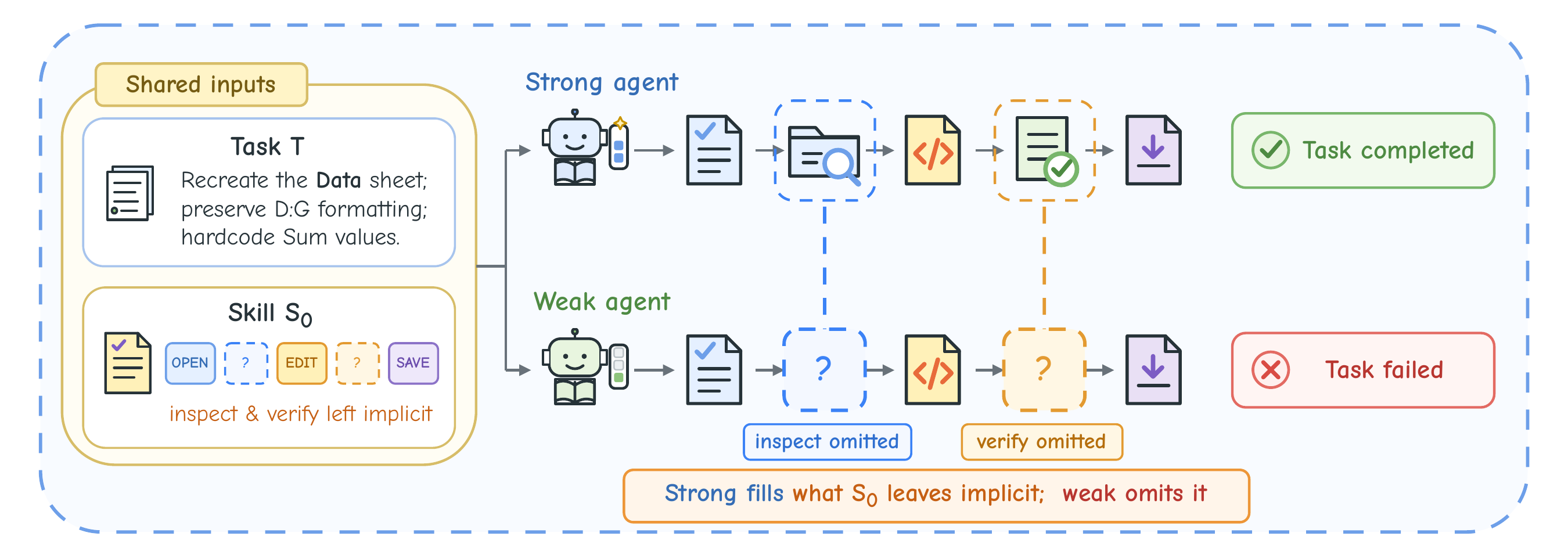}
    \caption{Motivation for Skill Transfer. Given the same task and skill $\mathcal{S}_0$, the strong agent supplies inspection and verification steps that the skill leaves implicit and completes the task. The weak agent omits those same steps and fails. This gap motivates Skill Transfer.}
    \label{fig:motivation}
\end{figure*}

% P2: problem with skills, strong vs. weak
However, \citet{skillsbench} report that the same curated skills yield various levels of gains across models and harnesses, demonstrating that some agents improve substantially while others gain little. In addition, \citet{skill-is-not-one-size-fit-all} show that skills of different granularity suit different agents, and a skill that benefits one agent can even harm another. Furthermore, we find a sharper form of this mismatch. Given one fixed skill, a strong and a weak agent not only differ in task success rate but also \textbf{follow markedly different procedures}, a divergence we quantify with graph edit distance (GED) over their execution trajectories~\citep{sanfeliu1983distance,gao2010survey}. Figure~\ref{fig:motivation} illustrates this mismatch. The skill itself is effective for the strong agent. The weak agent leaves much of its value unrealized because it omits procedural steps that the strong agent supplies implicitly. This raises a natural question:
% \why{Here I made it an explicit question, which should also be the core of our story. Not sure if the wording is correct.}

\begin{quote}
\emph{Can the full value of the skill be recovered for the weak agent without retraining it?}    
\end{quote}

% Specifically, the weak agent leaves much of the skill's value unrealized, not because the skill is deficient, but because it omits procedural steps that the strong agent supplies implicitly \why{"not because... but because..." looks too AI. not sure how to improve now.}.

% P3: formalize the problem "Skill Transfer".
% \why{If I were to write this paragraph, I would end the P2 with a question such as ``Given the misalignment, can we transfer the skill from strong to weak?" Then we can start this paragraph with a natural transition.}
We formalize the problem of closing this gap as \emph{Skill Transfer}. Specifically, given a skill that is already effective for a strong agent, we modify it so that a specific weak agent can use it effectively. We realize Skill Transfer by abstracting the strong and weak agents' trajectories into a shared structural representation, localizing their procedural gaps, and iteratively patching the skill to close them.

% We formalize the problem of closing this gap as \emph{Skill Transfer}: given a skill that is already sufficient for a strong agent, modify it so that a specific weak agent can use it effectively. We realize Skill Transfer by comparing the trajectories of the strong and weak agents on the same original skill, localizing the procedural gaps between them, and modifying the skill to close those gaps. \yf{Modify}
% Therefore, we propose Skill Transfer, which means transfer the skill to weak agents through the comparison of the trajectories between strong and weak agent using the same original skill. 

% P4: should introduce what we do in a little bit more detail.
% \why{Avoid defensive writing. "deliberately distinct from..." looks like a rebuttal. Also, avoid too many ``:'' and ``---''.}
% Skill Transfer is distinct from two adjacent lines of work. First, \emph{skill evolution}~\citep{voyager,evoskill,trace2skill,skillgrad} improves a skill so that it solves \emph{more} problems; its objective is the skill's absolute, model-agnostic capability, with no target executor in mind. Second, \emph{model distillation} changes the model's weights, from its classical form~\citep{hinton} to recent agent distillation that fine-tunes small models on teacher trajectories~\citep{agentdistill}. Skill Transfer changes neither: it re-expresses a fixed competence in a shared skill so that a weaker agent can execute it.
This problem matters for deployment. Real-world systems increasingly rely on smaller, cheaper agents to meet cost, latency, and resource constraints~\citep{edge-llm-review,ondevice-review}, whereas the strongest models remain expensive and resource-intensive. Skill Transfer lets such deployments reuse the large and growing corpus of skills authored for frontier models, rather than discarding them or retraining the model.

% P5: typically after introducing problem, motivation, our method, we introduce empirical results in this paragraph, followed by "contributions are three-fold:...".

Our contributions are as follows:
\begin{itemize}
    \item We formalize \emph{Skill Transfer}, the problem of adapting a skill that already works well for a strong agent so that a weak agent can use it effectively.
    \item We propose a trajectory-comparison method that abstracts strong and weak trajectories into typed execution structures, localizes their structural gaps, and repairs the skill with a diagnoser--patcher agent pair.
    \item We show across multiple benchmarks and agents that Skill Transfer substantially closes the strong--weak gap without modifying the weak agent.
\end{itemize}

\section{Related Work}

\paragraph{Memory Learning.}
A line of work improves an agent without weight updates by accumulating experience in an external memory: storing and retrieving observations, reflections, or long-term records~\citep{generativeagents,memgpt,memorybank,amem,mem0}, and distilling past trajectories into reusable, transferable insights~\citep{expel,reflexion,memcollab,preflect}. Such memory is auxiliary experience gathered at test time; in contrast, a skill is an authored, self-contained procedure that an agent loads on demand, and our work transfers this skill to a target agent rather than growing an experience store.

% There have been some works grow or optimize a skill library so that an agent solves more tasks---accumulating executable skills through self-practice~\citep{voyager,skillweaver}, discovering and verifying them from execution failures~\citep{evoskill,coevoskills,trace2skill,skillgrad}, or refining them via reinforcement learning~\citep{skillrl}. 
% MASA also notices that skill should be adapted to agents with different backbones~\citep{skill-is-not-one-size-fit-all}, but it just simply adds a model card (contains some metadata of the model) to the skill evolution pipeline to adapt to this model. However, the information of the model card is limited and unavailable for the closed-source model, making it unpractical.
% In contrast, we transfer a fixed, already-sufficient skill from a strong to a weak agent by explicitly localizing and closing the procedural gap between their trajectories with the same skill.
\paragraph{Skill Evolution.}
Several works grow or optimize a skill library so that an agent solves more tasks: accumulating executable skills through self-practice~\citep{voyager,skillweaver}, discovering and refining them from execution feedback~\citep{evoskill,coevoskills,trace2skill,skillgrad}, or evolving them via reinforcement learning~\citep{skillrl}. 
These methods raise a skill's absolute capability to solve more tasks, and do not distinguish the agent a skill was written for from the agent that has to run it. We instead adapt an existing skill effective on the strong agent to a weak agent, taking the gap between strong and weak trajectories as the signal. MASA~\citep{skill-is-not-one-size-fit-all} also claims that a skill should be adapted to different backbones, but conditions skill evolution on a pre-specified model card, a coarse profile that is largely unavailable for closed-source models.

% MASA also observes that a skill should be adapted to different backbones~\citep{skill-is-not-one-size-fit-all}: it conditions skill evolution on a model card (architecture metadata together with training and capability information) and searches for skill rewrites guided by that profile. This adaptation signal is a coarse, pre-specified description of the backbone, and much of this information is unavailable for frequently used closed-source models. In contrast, we require no model profile: we recover the concrete procedural gap directly by comparing the strong and weak agents' trajectories on the same skill and closing the divergence between them.
\paragraph{Agent Distillation.}
Distillation is a complementary route to transfer capability into a smaller model's parameters---from classical output~\citep{hinton,distillstep} to fine-tuning smaller LLMs on teacher-generated agent trajectories for tool use and reasoning~\citep{agenttuning,fireact,agentdistill}. All of these modify the model's weights and require training, whereas Skill Transfer is training-free, freezing the weak model and editing only the shared, human-readable skill. In this sense, Skill Transfer can be viewed as a form of training-free distillation: it distills the strong agent's procedural competence into the transferred skill rather than into model weights.

\section{Method}
\label{sec:method}

\begin{figure*}
    \centering
    \includegraphics[width=1\linewidth]{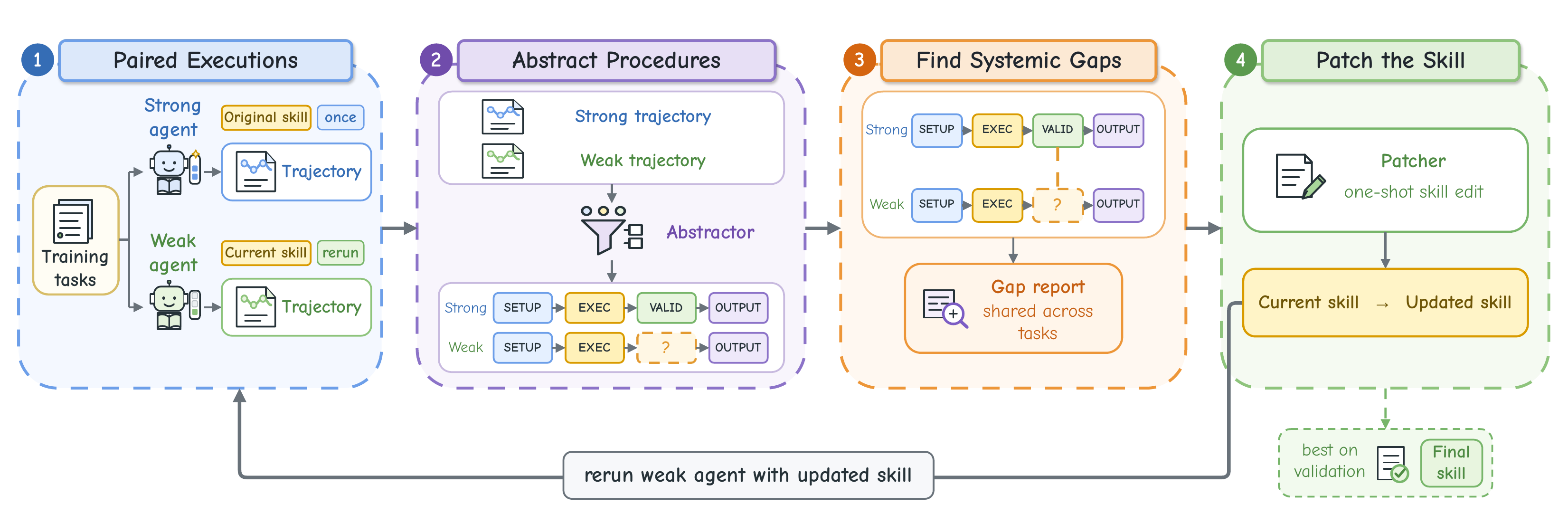}
    \caption{Skill Transfer overview. (1) The strong agent runs once under the original skill $\mathcal{S}_0$, while the weak agent reruns each iteration under the current skill $\mathcal{S}_k$ on the same training tasks. (2) The Abstractor turns both trajectories into typed node sequences (\S\ref{sec:abstraction}). (3) Aligning the paired structures localizes the gaps on each task and synthesizes the systemic ones into a gap report (\S\ref{sec:localization}--\S\ref{sec:diagnoser}). (4) The patcher edits only the skill, in one pass, to obtain $\mathcal{S}_{k+1}$, which is used to rerun the weak agent. After the iterations, $\mathcal{S}^\star$ is selected on $\mathcal{V}$.}
    \label{fig:pipeline}
\end{figure*}

% In this section, we introduce the Skill Transfer method.
Skill Transfer proceeds in four steps. It first abstracts weak and strong trajectories into typed execution structures (\S\ref{sec:abstraction}), then localizes where the weak procedure structurally diverges from the strong one on each task (\S\ref{sec:localization}), then synthesizes the systemic gaps across the whole training batch (\S\ref{sec:diagnoser}), and finally repairs the skill by turning each diagnosed gap into a targeted edit (\S\ref{sec:patcher}). \S\ref{sec:loop} states which agent carries out each step and how the steps iterate until the skill converges. Figure~\ref{fig:pipeline} gives an overview.

\subsection{Problem Formulation}
\label{sec:problem}
Let $\pi_s$ and $\pi_w$ denote a strong (teacher) and a weak (student) agent that share the same agentic configurations and, at inference time, the same skill; they differ only in their backbone model. A task is a pair $(t,y)$ of a task specification and its ground-truth answer, drawn from a distribution $\mathcal{D}$ over the target domain. Running an agent $\pi$ equipped with skill $\mathcal{S}$ on $t$ produces a trajectory $\tau$, a sequence of thoughts, tool calls, and observations, together with a final output $o$. A task metric scores the output against the ground truth, $R(o,y)\in[0,1]$, with $R{=}1$ for a fully correct output. The two play different roles below: $R(o,y)$ alone decides whether a task is solved, while $\tau$ is what we abstract and compare (\S\ref{sec:abstraction}--\S\ref{sec:localization}).  Let $\mathcal{S}_0$ be an original skill that is already effective for the strong agent. We define the objective of Skill Transfer as the weak agent's expected success over the domain,
\begin{equation}
J(\mathcal{S}) = \mathbb{E}_{(t,y)\sim\mathcal{D},\ (\tau,o)\sim\pi_w(t;\mathcal{S})}\big[\,R(o,y)\,\big],
\label{eq:objective}
\end{equation}
and adapt $\mathcal{S}_0$ to maximize $J(\mathcal{S})$ while leaving the weak model's parameters and the skill's intended capability unchanged. Because a skill is free-form text and $\mathcal{D}$ is unknown, we cannot maximize $J$ directly. Instead, from a small set of training tasks $\mathcal{T}=\{t_1,\dots,t_N\}$ that the strong agent solves, an iterative diagnoser-patcher loop proposes a sequence of candidate skills $\mathcal{S}_0,\dots,\mathcal{S}_K$, from which we select the adapted skill $\mathcal{S}^\star$ on a held-out validation set.

\subsection{Execution Structure Abstraction}
\label{sec:abstraction}

Raw trajectories are too noisy to compare directly, especially across different agents, because they use different tool syntax, follow different reasoning styles, and present the same step in different surface forms.
% \why{A little bit unclear. I understand backbones here as the LLMs. They have different tool syntax but the harness will handle them. Same with ``different surface forms''. Btw, avoid using ``:''}. 
We instead compare a typed intermediate representation, the \emph{execution structure}, obtained by abstracting each trajectory with an LLM (i.e., the \emph{Abstractor}).

Given a task $t$ and a trajectory $\tau$, the Abstractor emits an ordered sequence of \emph{procedure nodes} $X(v_1,\dots,v_m)$, one per distinct procedural step (Figure~\ref{fig:pipeline}b). Each node $v_i$ is a triple
\begin{equation}
v_i = \big(c^1_i,\, c^2_i,\, d_i\big).
\label{eq:node-tuple}
\end{equation}
% \why{here the explanations of the four annotations are too long. So we can avoid using "where xxx means yyy". Instead, we can do "Each node $v_i$ is a tuple (...). Here, $c_i^1$ means xxx ......} 
% where $c^1_i$ is a \emph{tier-1} label from a small, fixed, domain-agnostic set of six procedural roles: \textsc{setup} (preparation), \textsc{execute} (the primary domain action), \textsc{validate} (a correctness check before committing), \textsc{recover} (responding to failure), \textsc{output} (persisting a result), and \textsc{other} (anything fitting none of the above); $c^2_i$ is a \emph{tier-2} label, a short free-text tag the Abstractor invents to name the specific step (e.g. \texttt{borderless\_table\_fallback}, \texttt{formula\_write\_or\_repair}). Tier-1 gives a fixed vocabulary for cross-domain alignment; tier-2 carries the domain detail a patch needs. $\phi_i$ is a completion criterion, a concrete statement of when the step counts as done. $\iota_i$ is its evidence: the tool calls that realize the step, each with a code snippet and an observation spliced in verbatim from the raw trajectory, and a success/error/partial outcome. For skill-activation calls, $\iota_i$ also carries the agent's own stated reason for that choice, used in Section \ref{sec:localization} and Section \ref{sec:patcher} to fix wrong-skill routing.

Here, $c^1_i$ denotes a \emph{tier-1} label from a small, fixed, domain-agnostic set of six procedural roles: \textsc{setup} (preparation), \textsc{execute} (the primary domain action), \textsc{validate} (a correctness check before committing), \textsc{recover} (responding to failure), \textsc{output} (persisting a result), and \textsc{other} (anything fitting none of the above). $c^2_i$ denotes a \emph{tier-2} label, a short free-text tag the Abstractor invents to name the specific step (e.g. \texttt{borderless\_table\_fallback}, \texttt{formula\_write\_or\_repair}). Tier-1 gives a fixed vocabulary for cross-domain alignment, and tier-2 carries the domain detail a patch needs. $d_i$ holds the step's supporting detail: its completion criterion, the tool calls, code snippets, and observations that realize it, spliced in verbatim from the trajectory, and, when the step activates a skill, the agent's stated reason for that choice. Only the two labels enter the GED calculation of \S\ref{sec:localization}, but $d_i$ is what the gap analysis there reads and cites. Appendix~\ref{sec:appendix-structure} shows a complete execution structure.

We abstract a strong and a weak trajectory jointly in one pass so equivalent steps get the same tier-2 label; abstracting them separately lets the same technique drift to different labels and breaks the gap analysis below. 

\subsection{Structural Gap Localization}
\label{sec:localization}

\begin{figure*}[t]
    \centering
    \includegraphics[width=1\linewidth]{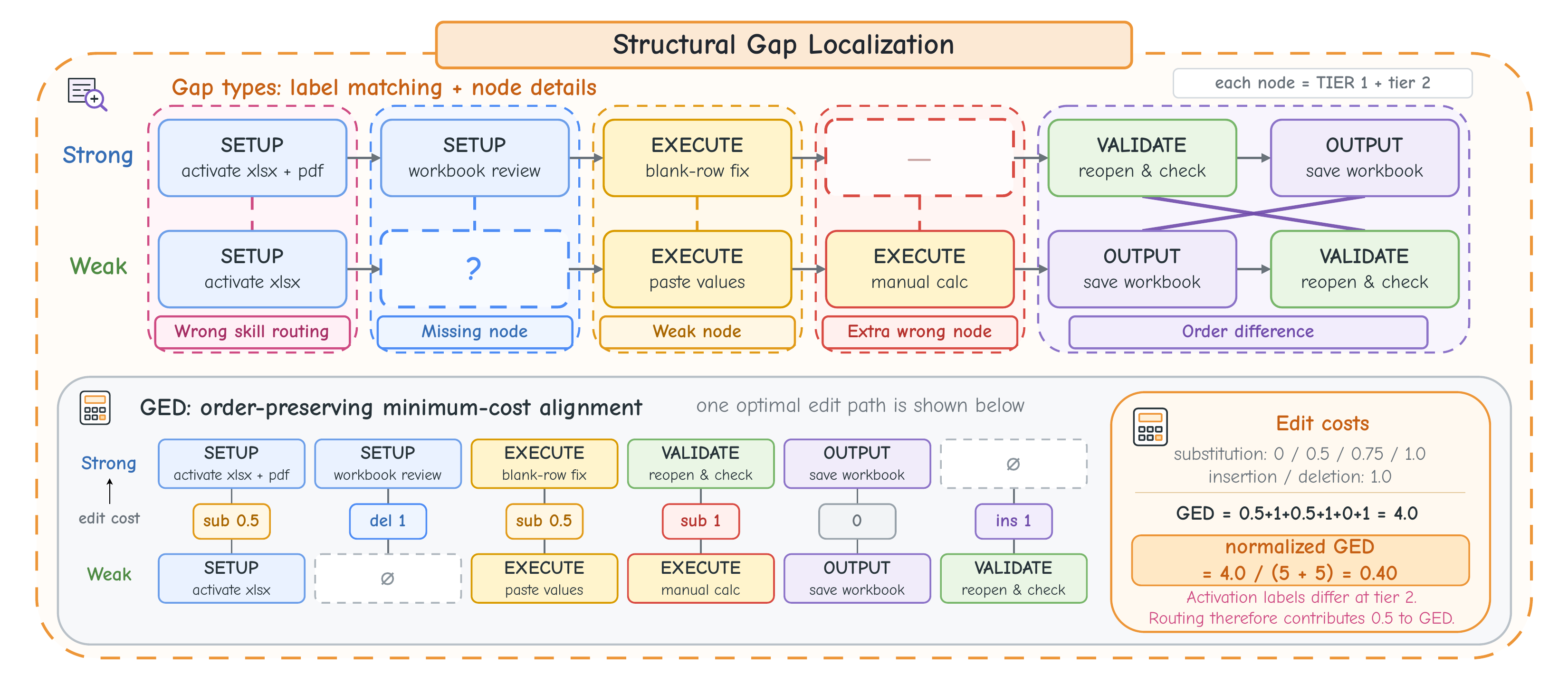}
    \caption{Structural gap localization on one task. \emph{Top}: matching the strong and weak node sequences by their tier-1/tier-2 labels surfaces the five gap types; \texttt{wrong\_skill\_routing} is read from the activated skills recorded in $d$. \emph{Bottom}: one minimum-cost alignment of the same pair, charging $\delta$ (Eq.~\ref{eq:subcost}) per matched pair and one per unmatched node, which gives $\mathrm{GED}=4.0$ and $\widehat{\mathrm{GED}}=4.0/(5{+}5)=0.40$.}
    \label{fig:localization}
\end{figure*}

Given strong and weak structures $X_s=(v_1,\dots,v_m)$ and $X_w=(v'_1,\dots,v'_n)$ for the same task, we first quantify their divergence through graph edit distance (ged), and then localize it to specific steps. Figure~\ref{fig:localization} shows an example.

\paragraph{Graph edit distance.} 
We treat each execution structure as a path graph ordered by execution time and measure dissimilarity with graph edit distance~\citep{sanfeliu1983distance,gao2010survey}, computed by the standard sequence-alignment dynamic program with a substitution cost $\delta(v,v')$ graded on the two-level category rather than binary match/mismatch:
\begin{equation}
\delta(v,v') =
\begin{cases}
0.00 & c^1{=}c'^1 \wedge c^2{=}c'^2\\
0.50 & c^1{=}c'^1 \wedge c^2{\neq}c'^2\\
0.75 & c^1{\neq}c'^1 \wedge c^2{=}c'^2\\
1.00 & c^1{\neq}c'^1 \wedge c^2{\neq}c'^2
\end{cases}
\label{eq:subcost}
\end{equation}
and an insertion/deletion cost of $1.0$. An alignment $\mathcal{A}$ pairs each node of $X_s$ with at most one node of $X_w$, and preserves order: if one step of $X_s$ precedes another, their partners in $X_w$ appear in the same order. GED is the cost of the cheapest alignment,
\begin{equation}
\mathrm{GED}(X_s,X_w) = \min_{\mathcal{A}} \Big[\, \sum_{(i,j)\in\mathcal{A}} \delta(v_i,v'_j) \;+\; \big(m-|\mathcal{A}|\big) \;+\; \big(n-|\mathcal{A}|\big) \Big],
\label{eq:ged}
\end{equation}
where the last two terms charge one unit for every node left unmatched, deleted from $X_s$ or inserted into $X_w$; the minimum is found by the standard $O(mn)$ dynamic program. Matching nothing costs $|X_s|+|X_w| = m+n$, the worst case and hence the normalizer:
\begin{equation}
\widehat{\mathrm{GED}}(X_s,X_w) = \frac{\mathrm{GED}(X_s,X_w)}{m+n} \in [0,1].
\label{eq:ged-norm}
\end{equation}
GED is a single scalar summarizing how much the two procedures disagree overall; it is a diagnostic signal, not a success criterion, so a task is scored correct from its output alone.

\paragraph{Gap type.} A single scalar cannot say where the two procedures disagree, so we also localize divergence to individual steps. We write $C(X)$ for the multiset of tier-1 roles occurring in $X$, so that a role performed twice counts twice, $K(X)$ for the skills activated in $X$ (read off the activation nodes' $d$), and $\mathcal{M}$ for the label matching that pairs $v_i$ with $v'_j$ when $(c^1_i,c^2_i)=(c'^1_j,c'^2_j)$, falling back to tier-1 alone when no tier-2 counterpart exists. Unlike the alignment $\mathcal{A}$ of Eq.~\ref{eq:ged}, $\mathcal{M}$ need not preserve order, which is what makes step reordering visible. Comparing matched and unmatched nodes yields five gap types (Figure~\ref{fig:pipeline}c):

\begin{itemize}[leftmargin=*,itemsep=1pt,topsep=2pt]
    \item \textbf{missing\_node}: $C(X_s)\setminus C(X_w)\neq\emptyset$, as a multiset difference. The weak run performs a tier-1 role fewer times than the strong run, either never at all (e.g. no \textsc{validate} step) or fewer times than needed when the role recurs.
     \item \textbf{weak\_node}: some $(i,j)\in\mathcal{M}$ has $d'_j\prec d_i$. The role is present but executed to a lower bar: a looser completion criterion, fewer artifacts, a worse technique, or a correct tool with an incorrect computed result.
    \item \textbf{extra\_wrong\_node}: some $j$ is unmatched by $\mathcal{M}$. The weak run takes a step with no counterpart in $X_s$ that fails.
    \item \textbf{order\_difference}: some $(i,j),(i',j')\in\mathcal{M}$ have $i<i'$ but $j>j'$. Both runs contain the two steps, in opposite relative order.
    \item \textbf{wrong\_skill\_routing}: $K(X_s)\setminus K(X_w)\neq\emptyset$. When a task may draw on more than one skill, the weak agent's activated set is missing one the strong agent used.
\end{itemize}

The ordering $\prec$ on step is judged from the cited $d$, and every reported gap must quote the specific code snippet or observation behind it, falling back to the raw trajectory only when the structures are insufficient.

\subsection{Multi-Case Gap Synthesis} 
\label{sec:diagnoser}
% \yf{make it a step}
% To find the systemic gaps between the strong and weak agents,  we diagnose gaps over the whole training batch $\mathcal{T}$ at once rather than one task at a time.
% At iteration $k$, the diagnoser receives the current skill $\mathcal{S}_k$, the fixed strong structures $\{X_s^{(t)}\}_{t\in\mathcal{T}}$, the current-iteration weak structures $\{X_w^{(t)}\}_{t\in\mathcal{T}}$, a per-task validator report (a finer-grained comparison of $\pi_w$'s output against $\pi_s$'s than the binary $y$, as presented in Appendix~\ref{sec:appendix-validator}), and (for $k{>}0$) the gap reports and patch history of all prior iterations $0,\dots,k{-}1$.

\S\ref{sec:localization} types the gaps on a single task. However, a skill edit that chases one task will not generalize, so this step applies that typing to every task in the training batch $\mathcal{T}$ and synthesizes the results. Besides the paired structures $\{X_s^{(t)}\},\{X_w^{(t)}\}$ and the current skill, it reads a per-task validator report, a finer-grained comparison of $\pi_w$'s output against $\pi_s$'s than $R$ (Appendix~\ref{sec:appendix-validator}), as well as earlier iterations' gap reports and patch history.

Per-task gaps describing the same weakness are grouped across tasks into a set of systemic gaps $G=\{g_1,g_2,\dots\}$. Each $g\in G$ carries a gap type, metadata (label, description), and two case sets: $F_g\subseteq\mathcal{T}$, the tasks where $\pi_w$ failed with the gap present, and $P_g\subseteq\mathcal{T}$, the tasks where $\pi_w$ passed despite it. A gap is admitted for repair if $|F_g|\geq 1$ or $|P_g|\geq 2$. The second condition catches a gap that has not yet caused a failure but recurs in runs that passed anyway, which we treat as a latent robustness risk rather than a non-issue.

\paragraph{Strategy selection and patch hint.} Each gap admitted for repair is assigned some of the six mutually-exclusive strategies. \textsc{mandate} adds a new required workflow step with an explicit completion criterion; \textsc{guidance} sharpens an existing but under-specified step; \textsc{prohibition} forbids a specific action the weak agent takes that the strong agent never does; \textsc{hard-stop} forces the agent to detect and discard a wrong result rather than report success on it; \textsc{reorder} fixes step sequencing; and \textsc{description\_patch} extends a skill's description (rather than its main content) using both agents' stated reasons for choosing a skill, to fix wrong-skill routing. Table~\ref{tab:strategy} maps each gap type to the strategies that repair it; when the gap has been seen in an earlier iteration, its status further constrains the choice (next paragraph and Table~\ref{tab:status}). The chosen strategy is written into a patch hint, free-form prose covering four elements: \emph{target}, the file and section to edit; \emph{strategy}, the chosen label; \emph{wording}, the instruction text to install, which must be phrased generally enough to cover cases beyond the ones cited; and, when a previous iteration already patched this gap, a critique of why that patch failed to close it. This hint is the primary handoff to the repair step (\S\ref{sec:patcher}), which executes it without redoing the diagnosis. Appendix~\ref{sec:appendix-gapreport} shows one gap entry with its hint.

\begin{table*}[t]
\centering\small
\begin{tabular}{@{}p{3.2cm}p{10.15cm}@{}}
\toprule
\textbf{Gap type} & \textbf{Strategy} \\
\midrule
missing\_node & \textsc{mandate} (+ \textsc{prohibition} if the tempting wrong path is non-obvious) \\
weak\_node & \textsc{guidance} / \textsc{prohibition} / \textsc{hard-stop} / \textsc{mandate}, chosen by root cause \\
extra\_wrong\_node & \textsc{prohibition} \\
order\_difference & \textsc{reorder} \\
wrong\_skill\_routing & \textsc{description\_patch} \\
\bottomrule
\end{tabular}
\caption{Gap type $\to$ patch strategy mapping written into the patch hint.}
\label{tab:strategy}
\end{table*}

\paragraph{Status.} At iteration $k{=}0$ every gap is trivially NEW, since no prior report exists to match against. From $k{\geq}1$ on, each current gap is instead matched to its counterpart in iteration $k{-}1$'s report by label, and classified by the change in its failing-case count $\Delta=|F_g^{(k)}|-|F_g^{(k-1)}|$. Table~\ref{tab:status} shows the resulting categories and how each overrides the default strategy choice. Eligible gaps are then ranked for repair by status tier (REGRESSION $>$ PERSISTS $>$ NEW $>$ IMPROVEMENT), then by failing-case count descending within a tier.

\begin{table*}[t]
\centering\small
\begin{tabular}{@{}p{2.4cm}p{3.4cm}p{7.2cm}@{}}
\toprule
\textbf{Status} & \textbf{Trigger} & \textbf{Effect on the patch hint} \\
\midrule
NEW & no iteration-$(k{-}1)$ match & take the strategy from Table~\ref{tab:strategy}; no prior-patch critique \\
RESOLVED & $|F_g|{=}0$ and $|P_g|<2$ & drop the gap; emit no patch \\
IMPROVEMENT & $\Delta<0$ & keep the previous strategy and supplement it for the cases still failing \\
PERSISTS & $\Delta=0$ & \textsc{guidance} is forbidden; escalate to a different mechanism and \emph{replace} the standing instruction rather than append to it \\
REGRESSION & $\Delta>0$ & revert or rescope the harmful edit first, then re-derive the strategy from the root cause \\
\bottomrule
\end{tabular}
\caption{Cross-iteration gap status and how it overrides the strategy of Table~\ref{tab:strategy}.}
\label{tab:status}
\end{table*}

\subsection{One-Shot Skill Patching}
\label{sec:patcher}

Given the ranked gap report of \S\ref{sec:diagnoser}, this step applies the edits it describes in a single consolidated pass:
\begin{itemize}[leftmargin=*,itemsep=1pt,topsep=2pt]
    \item \textbf{Read and group}: read every admitted gap's patch hint and group hints by the target file section they name.
    \item \textbf{Synthesize}: for each affected section, design one coherent edit satisfying every hint touching it.
    % \item \textbf{Apply}: commit all sections in one batched multi-section edit (one exact old-text/new-text replacement per section; never a full-file rewrite).
    \item \textbf{Apply}: replace the old text that needs editing in all sections in one multi-section edit.
    \item \textbf{Verify}: re-read the patched skill, confirm every gap in the repair queue has been fixed and no section contradicts another.
\end{itemize}

\subsection{Iterative Loop}
\label{sec:loop}

Two LLM agents carry out the steps above. A \emph{diagnoser} performs localization and synthesis (\S\ref{sec:localization}--\S\ref{sec:diagnoser}), reading the paired structures and emitting the ranked gap report; a \emph{patcher} performs the repair (\S\ref{sec:patcher}), and is the only one of the two that may edit the skill. Skill Transfer alternates the two over the fixed batch $\mathcal{T}$. At iteration $0$, $\mathcal{T}$ is restricted once to tasks $\pi_s$ solves within up to $M$ independent attempts (excluding tasks the strong agent still fails after $M$ tries) and then held fixed for every later iteration. In each iteration, $\pi_w$ reruns under $\mathcal{S}_k$ on $\mathcal{T}$ first, the diagnoser then produces the gap report, and the patcher turns $\mathcal{S}_k$ into $\mathcal{S}_{k+1}$. The loop stops at a fixed iteration budget $K{=}$\,\texttt{max\_edits}, or earlier if the skill is left unchanged for two consecutive iterations. We take the adapted skill $\mathcal{S}^\star$ to be the iterate with the highest pass rate on a held-out validation set $\mathcal{V}$,
\begin{equation}
\mathcal{S}^\star = \arg\max_{k\in\{0,\dots,K\}} \mathrm{acc}_{\mathcal{V}}(\mathcal{S}_k).
\label{eq:sstar}
\end{equation}
Algorithm~\ref{alg:skilltransfer} summarizes the full loop.

\begin{algorithm}[t]
\small
\caption{Skill Transfer}
\label{alg:skilltransfer}
\begin{algorithmic}[1]
\Require original skill $\mathcal{S}_0$, already effective for the strong agent $\pi_s$; weak agent $\pi_w$; candidate training tasks $\mathcal{T}_0$; validation set $\mathcal{V}$; iteration budget $K$; strong-agent attempts $M$
\Ensure a skill adapted to $\pi_w$
\Statex \hrulefill
\Statex \emph{Strong-agent reference, run once}
\State Run $\pi_s$ on every task in $\mathcal{T}_0$ under $\mathcal{S}_0$, retrying a task up to $M$ times
\State $\mathcal{T} \gets$ the tasks $\pi_s$ solved \Comment{the rest offer no reference procedure}
\Statex \emph{Adaptation loop, with the weak agent rerun every iteration}
\State $\mathcal{S} \gets \mathcal{S}_0$; \ $\mathcal{S}^\star \gets \mathcal{S}_0$ \Comment{current and best-so-far skill}
\For{$k = 1$ \textbf{to} $K$}
  \State Run $\pi_w$ on $\mathcal{T}$ under $\mathcal{S}$
  \State Abstract both strong and weak trajectories jointly \Comment{\S\ref{sec:abstraction}}
  \If{$\pi_w$ solved every task in $\mathcal{T}$}
    \State \textbf{break} \Comment{no gap left to close}
  \EndIf
  \State Localize the gaps in each strong--weak structure pair \Comment{\S\ref{sec:localization}}
  \State Group the per-task gaps into systemic ones; rank them and write one patch hint each \Comment{\S\ref{sec:diagnoser}}
  \State $\mathcal{S} \gets$ apply all patch hints to $\mathcal{S}$ in a single pass \Comment{\S\ref{sec:patcher}}
  \If{$\mathcal{S}$ scores higher than $\mathcal{S}^\star$ on $\mathcal{V}$}
    \State $\mathcal{S}^\star \gets \mathcal{S}$
  \EndIf
  \If{$\mathcal{S}$ has been left unchanged in two consecutive iterations}
    \State \textbf{break} \Comment{converged}
  \EndIf
\EndFor
\State \Return $\mathcal{S}^\star$
\end{algorithmic}
\end{algorithm}

\section{Experiments}
\label{sec:experiments}

\subsection{Setup}
\label{sec:setup}

\paragraph{Data.} We evaluate on three benchmarks, each with 20 and 30 further cases for training and validation. SpreadsheetBench~\citep{ma2024spreadsheetbench} collects real-world spreadsheet-manipulation requests from online forums; each task gives the agent an input workbook and an instruction (e.g. insert a formula, reformat a range, filter rows); we use a 200-case test split. OfficeQA~\citep{officeqa} poses grounded numerical-reasoning questions over historical U.S. Treasury Bulletin documents, requiring the agent to locate the relevant table across long, noisy source text and compute a numeric answer; we use a 70-case test split. DABench~\citep{dabench} poses data-analysis questions over tabular CSV files, requiring the agent to write and execute analysis code and report a closed-form answer; we use a 150-case test split.

\paragraph{Agents.} We run the experiments under two model families. For GPT, the backbone of strong agent $\pi_s$ is \texttt{gpt-5.4}; weak agents $\pi_w$ use \texttt{gpt-5.4-mini} and \texttt{gpt-4.1-mini}. For Qwen3.5, we run SpreadsheetBench and DABench with \texttt{Qwen3.5-122B-A10B-FP8} as $\pi_s$ and \texttt{Qwen3.5-4B} (SpreadsheetBench) or \texttt{Qwen3.5-2B} (DABench) as $\pi_w$, with $\pi_s$ and $\pi_w$ each served on a single H200 GPU; we switch to the smaller \texttt{Qwen3.5-2B} on DABench because \texttt{Qwen3.5-4B} was already too close to $\pi_s$ to leave a meaningful gap to close. All agents are implemented as a tool-using agent loop on the OpenAI Agents SDK~\citep{openai-agents-sdk}. For the GPT experiments, the Abstractor, Diagnoser, and Patcher run on \texttt{gpt-5.4} and the Validator on \texttt{gpt-5.4-mini}; for Qwen3.5, all four run on \texttt{Qwen3.5-122B-A10B-FP8}.

\paragraph{Skills and baselines.} Each benchmark starts from an original skill $\mathcal{S}_0$ that is already effective for $\pi_s$, reused unchanged across both model families. For SpreadsheetBench, $\mathcal{S}_0$ is Anthropic's official xlsx skill~\citep{anthropic_skills}. For OfficeQA, $\mathcal{S}_0$ is a pair of third-party public skills finance~\citep{finance-skill} and ripgrep~\citep{ripgrep-skill}. For DABench, $\mathcal{S}_0$ is a pair of third-party public skills data-scientist~\citep{datascientist-skill} and statistical-analysis~\citep{statistical-analysis-skill}. We compare four methods that all start from the same $\mathcal{S}_0$: \emph{no skill}, running without any skill; \emph{original skill}, the unmodified $\mathcal{S}_0$; \emph{EvoSkill}~\citep{evoskill}, a skill-evolution baseline that optimizes the skill's capability; and our method Skill Transfer, run for an iteration budget of $K{=}6$ with up to $M{=}5$ strong-agent attempts (Algorithm~\ref{alg:skilltransfer}).

\paragraph{Metrics.} Beyond task pass rate, we report benchmark-specific fine-grained metrics (cell match ratio for SpreadsheetBench, accuracy within a relative tolerance of $\leq$0.1\%/$\leq$1\%/$\leq$5\% for OfficeQA, and DABench's own accuracy-by-sub-question / accuracy-proportional-by-sub-question for DABench), and average $\widehat{\mathrm{GED}}$ (Eq.~\ref{eq:ged-norm}) to measure how procedurally close the weak agent's run is to the strong agent's.

\subsection{Main Results}
\label{sec:mainresults}

\begin{table*}[t]
\centering\footnotesize
\setlength{\tabcolsep}{2.0pt}
\begin{tabular}{@{}ll ccc ccccc cccc@{}}
\toprule
& & \multicolumn{3}{c}{\textbf{SpreadsheetBench}} & \multicolumn{5}{c}{\textbf{OfficeQA}} & \multicolumn{4}{c}{\textbf{DABench}} \\
\cmidrule(lr){3-5} \cmidrule(lr){6-10} \cmidrule(lr){11-14}
\textbf{Agent} & \textbf{Method} & \textbf{P\% $\uparrow$} & \textbf{CM $\uparrow$} & \textbf{G $\downarrow$} & \textbf{P\% $\uparrow$} & \textbf{$\leq$.1 $\uparrow$} & \textbf{$\leq$1 $\uparrow$} & \textbf{$\leq$5 $\uparrow$} & \textbf{G $\downarrow$} & \textbf{P\% $\uparrow$} & \textbf{ByQ $\uparrow$} & \textbf{PrQ $\uparrow$} & \textbf{G $\downarrow$} \\
\midrule
\multirow{2}{*}{\shortstack[l]{Strong\\\texttt{gpt-5.4}}} & No skill & 59.0 & 0.717 & -- & 52.9 & 57.1 & 62.9 & 67.1 & -- & 82.0 & 86.3 & 85.0 & -- \\
 & Original skill & 67.5 & 0.807 & -- & 57.1 & 57.1 & 67.1 & 72.9 & -- & 86.0 & 89.8 & 88.6 & -- \\
\midrule
\multirow{4}{*}{\shortstack[l]{Weak\\\texttt{gpt-4.1-mini}}} & No skill & 15.5 & 0.468 & 0.280 & \underline{17.1} & \underline{17.1} & \underline{24.3} & \underline{27.1} & 0.335 & 74.7 & 80.4 & 79.9 & 0.204 \\
 & Original skill & 25.5 & 0.486 & 0.203 & 8.6 & 12.9 & 15.7 & 15.7 & 0.297 & 68.7 & 72.9 & 74.3 & 0.218 \\
 & EvoSkill & \underline{36.5} & \underline{0.541} & \underline{0.187} & 14.3 & \underline{17.1} & \underline{24.3} & 25.7 & \underline{0.251} & \underline{79.3} & \underline{82.8} & \underline{83.2} & \underline{0.185} \\
 & Skill Transfer & \textbf{48.5} & \textbf{0.638} & \textbf{0.154} & \textbf{18.6} & \textbf{18.6} & \textbf{25.7} & \textbf{28.6} & \textbf{0.177} & \textbf{80.7} & \textbf{83.2} & \textbf{85.2} & \textbf{0.158} \\
\midrule
\multirow{4}{*}{\shortstack[l]{Weak\\\texttt{gpt-5.4-mini}}} & No skill & 48.0 & 0.662 & 0.227 & 21.4 & 24.3 & 25.7 & 30.0 & 0.275 & 77.3 & 85.2 & 84.3 & 0.243 \\
 & Original skill & 39.5 & 0.632 & 0.111 & 14.3 & 18.6 & 22.9 & 28.6 & 0.168 & 81.3 & 87.6 & 86.0 & 0.126 \\
 & EvoSkill & \underline{50.0} & \underline{0.683} & \underline{0.099} & \underline{27.1} & \underline{28.6} & \textbf{35.7} & \textbf{40.0} & \underline{0.127} & \textbf{86.0} & \underline{87.9} & \underline{87.2} & \textbf{0.083} \\
 & Skill Transfer & \textbf{51.5} & \textbf{0.691} & \textbf{0.095} & \textbf{30.0} & \textbf{31.4} & \textbf{35.7} & \underline{37.1} & \textbf{0.109} & \textbf{86.0} & \textbf{89.7} & \textbf{89.9} & \underline{0.109} \\
\bottomrule
\end{tabular}
\caption{SpreadsheetBench, OfficeQA, and DABench. Within each weak agent, \textbf{bold} is the best method for that column and \underline{underline} the second best. CM = Cell Match; G = $\widehat{\mathrm{GED}}$; $\leq X$ is accuracy within $X\%$ relative tolerance of the gold numeric answer; ByQ/PrQ are DABench's own accuracy-by-sub-question and accuracy-proportional-by-sub-question metrics.}
\label{tab:combined-results}
\end{table*}

Table~\ref{tab:combined-results} reports pass rate and the fine-grained metrics for the GPT family on all three benchmarks; Tables~\ref{tab:qwen-sb} and \ref{tab:qwen-dabench} report the Qwen3.5 family on SpreadsheetBench and DABench (we omit OfficeQA for Qwen3.5: its source documents routinely exceed Qwen3.5's 262{,}144-token context window, so the task is not fully well-defined for these backbones). Across all eight weak configurations, Skill Transfer achieves the best pass rate, beating the skill-evolution baseline EvoSkill by up to 12.0pp and the original skill by 4.7--23.0pp.
EvoSkill's proposer sees only the weak agent's own failures, never a strong-agent trajectory to diff against, so its edits add generic content rather than closing the specific gap the weak agent has.

The original skill's own effect on weak agents is not uniformly positive: it underperforms no skill in six of the eight weak configurations (e.g. 48.0$\to$39.5 for \texttt{gpt-5.4-mini} on SpreadsheetBench, 17.1$\to$8.6 for \texttt{gpt-4.1-mini} on OfficeQA), which shows that a skill written around the strong agent's implicit steps can actively mislead a weaker one that lacks them. Where the original skill does help, Skill Transfer still adds to it. On SpreadsheetBench \texttt{gpt-4.1-mini} goes from 15.5 with no skill to 25.5 with the original skill and 48.5 with Skill Transfer, closing 55\% of the distance the original skill leaves to \texttt{gpt-5.4}. On DABench \texttt{gpt-5.4-mini} goes 77.3 $\to$ 81.3 $\to$ 86.0 and closes that distance entirely, matching \texttt{gpt-5.4}'s pass rate and its ByQ/PrQ (89.7/89.9 against 89.8/88.6). Appendix~\ref{sec:appendix-skilldiff} shows the accumulated edits between an original skill and the skill Skill Transfer produced from it.

% We check whether skill transfer generalizes beyond the GPT family by repeating the same pipeline on Qwen3.5. Tables~\ref{tab:qwen-sb} and \ref{tab:qwen-dabench} show the same pattern as the GPT experiments. Skill Transfer is the best-performing method for both weak agents, beating EvoSkill and the original skill on both SpreadsheetBench and DABench. We omit OfficeQA for this model family: OfficeQA's source documents routinely exceed Qwen3.5's 262{,}144-token context window, so the task is not fully well-defined for these backbones.

\begin{table*}[t]
\centering
\begin{minipage}{0.45\textwidth}
\centering\footnotesize
\setlength{\tabcolsep}{2.5pt}
\begin{tabular}{@{}ll ccc@{}}
\toprule
\textbf{Agent} & \textbf{Method} & \textbf{P\% $\uparrow$} & \textbf{CM $\uparrow$} & \textbf{G $\downarrow$} \\
\midrule
\multirow{2}{*}{\shortstack[l]{Strong\\\texttt{Qwen3.5-122B}}} & No skill & 55.0 & 70.5 & -- \\
 & Original skill & 59.5 & 72.2 & -- \\
\midrule
\multirow{4}{*}{\shortstack[l]{Weak\\\texttt{Qwen3.5-4B}}} & No skill & 22.0 & 38.1 & 0.281 \\
 & Original skill & 15.5 & 35.5 & 0.222 \\
 & EvoSkill & 17.0 & 33.3 & 0.223 \\
 & Skill Transfer & \textbf{26.0} & \textbf{42.5} & \textbf{0.204} \\
\bottomrule
\end{tabular}
\captionof{table}{Qwen3.5 on SpreadsheetBench.}
\label{tab:qwen-sb}
\end{minipage}
\hfill
\begin{minipage}{0.50\textwidth}
\centering\footnotesize
\setlength{\tabcolsep}{2.2pt}
\begin{tabular}{@{}ll cccc@{}}
\toprule
\textbf{Agent} & \textbf{Method} & \textbf{P\% $\uparrow$} & \textbf{ByQ $\uparrow$} & \textbf{PrQ $\uparrow$} & \textbf{G $\downarrow$} \\
\midrule
\multirow{2}{*}{\shortstack[l]{Strong\\\texttt{Qwen3.5-122B}}} & No skill & 72.7 & 88.9 & 87.8 & -- \\
 & Original skill & 74.0 & 89.4 & 88.0 & -- \\
\midrule
\multirow{4}{*}{\shortstack[l]{Weak\\\texttt{Qwen3.5-2B}}} & No skill & 19.3 & 47.6 & 43.0 & 0.259 \\
 & Original skill & 7.3 & 48.8 & 45.8 & 0.273 \\
 & EvoSkill & 13.3 & 42.6 & 45.5 & 0.252 \\
 & Skill Transfer & \textbf{22.0} & \textbf{48.9} & \textbf{48.4} & \textbf{0.243} \\
\bottomrule
\end{tabular}
\captionof{table}{Qwen3.5 on DABench.}
\label{tab:qwen-dabench}
\end{minipage}
\end{table*}

\subsection{Structural Convergence}
\label{sec:gedresults}

$\widehat{\mathrm{GED}}$ decreases monotonically from no skill through original skill and EvoSkill to Skill Transfer in 4 of 8 configurations (both gpt-based weak agents on SpreadsheetBench and OfficeQA), and Skill Transfer reaches the lowest $\widehat{\mathrm{GED}}$ in 7 of 8. Where the pattern holds, it shows that Skill Transfer closes the procedural gap identified in \S\ref{sec:localization} rather than improving outputs incidentally. The weak agent has learned some of the procedural steps the strong agent performs.

\subsection{Skill Transfer Also Benefits the Strong Agent}
\label{sec:strongbenefits}

\begin{table*}[t]
\centering
\begin{minipage}{0.45\textwidth}
\centering\footnotesize
\setlength{\tabcolsep}{13pt}
\begin{tabular}{@{}llc@{}}
\toprule
\textbf{Method} & \textbf{Adapted for} & \textbf{Pass \% $\uparrow$} \\
\midrule
No skill & -- & 59.0 \\
Original skill & -- & 67.5 \\
Skill Transfer & \texttt{gpt-4.1-mini} & 76.5 \\
Skill Transfer & \texttt{gpt-5.4-mini} & 81.3 \\
\bottomrule
\end{tabular}
\captionof{table}{Strong agent (\texttt{gpt-5.4}) on SpreadsheetBench, using a skill adapted for a weak agent.}
\label{tab:strong-uses-adapted}
\end{minipage}
\hfill
\begin{minipage}{0.45\textwidth}
\centering\footnotesize
\setlength{\tabcolsep}{13pt}
\begin{tabular}{@{}llc@{}}
\toprule
\textbf{Weak agent} & \textbf{Skill} & \textbf{Acc \% $\uparrow$} \\
\midrule
\multirow{2}{*}{\texttt{gpt-4.1-mini}} & Original skill & 44.4 \\
 & Skill Transfer & \textbf{69.2} \\
\multirow{2}{*}{\texttt{gpt-5.4-mini}} & Original skill & 75.6 \\
 & Skill Transfer & \textbf{77.6} \\
\bottomrule
\end{tabular}
\captionof{table}{Out-of-distribution: WikiTableQuestions, using the xlsx skill adapted on SpreadsheetBench.}
\label{tab:wtq-ood}
\end{minipage}
\end{table*}

Although Skill Transfer targets the weak agent's procedural gap, Table~\ref{tab:strong-uses-adapted} shows the adapted skill also raises the strong agent's own pass rate on SpreadsheetBench, from 67.5\% to 76.5--81.3\% depending on which weak agent the skill was adapted for. A patch that makes an implicit step explicit does not only help a weak model that would skip the step totally; it also helps a stronger model that performs the step inconsistently. The adapted skill converts latent or inconsistent capability into more reliable behavior, benefiting the strong agent too.

\subsection{Out-of-Distribution Generalization}
\label{sec:oodresults}

We test whether a skill adapted on one benchmark transfers to a structurally related but distinct task: WikiTableQuestions~\citep{pasupat2015compositional} (WTQ), where the agent answers a natural-language question about a table by writing the result into a spreadsheet cell rather than editing a workbook. Table~\ref{tab:wtq-ood} reports results for both weak agents, using the xlsx skill adapted on SpreadsheetBench. For \texttt{gpt-4.1-mini}, accuracy on 250 held-out WTQ questions rises from 44.4\% to 69.2\% (+24.8pp). For \texttt{gpt-5.4-mini}, accuracy rises from 75.6\% to 77.6\% (+2.0pp) -- a much smaller gain, consistent with \texttt{gpt-5.4-mini}'s already-high accuracy under the original skill leaving less headroom to close. The skill was never trained on WTQ; the gains come from the same xlsx-handling procedure (e.g. correct cell addressing, formula evaluation, output placement) that Skill Transfer made explicit for SpreadsheetBench. The generalization demonstrates that Skill Transfer does not overfit on specific tasks, but truly closes the procedural gaps between weak and strong agents.

\section{Conclusion}
\label{sec:conclusion}

We formalized \emph{skill transfer}: given a skill already effective for a strong agent, transfer it to a specific weak agent without touching that agent's parameters. Our method abstracts both agents' trajectories into typed execution structures, localizes their gaps with a graded graph edit distance, and repairs the skill through an iterative diagnoser--patcher loop. Across three benchmarks and four weak agents from two model families, Skill Transfer raises pass rate over both the unmodified original skill and a skill-evolution baseline, and $\widehat{\mathrm{GED}}$ decreases toward the strong agent's structure in most configurations, so the gain comes from closing the procedural gap of \S\ref{sec:localization}. The adapted skill also improves the strong agent it was never adapted for, and transfers to a related task it was never trained on.
However, two conditions bound the method. First, it needs a strong agent and a small set of tasks that the agent solves, since the strong trajectories are the reference the diagnoser diffs against; without such a reference there is no gap to localize. Second, the gain depends on why the weak agent fails. Procedural failures are largely recoverable by making the missing steps explicit, whereas failures bound by capability the weak model does not have are difficult to recover from.
Within these bounds, skill transfer offers a training-free route to reusing the growing corpus of skills authored for frontier models on the smaller, cheaper agents that real deployments often require.

%%%%%%%%%%%%%%%%%%%%%%%%%%%%%%%%%%%%%%%%%%%%%%%%%%%%%%%%%%%%

\bibliographystyle{plainnat}
\bibliography{custom}

%%%%%%%%%%%%%%%%%%%%%%%%%%%%%%%%%%%%%%%%%%%%%%%%%%%%%%%%%%%%

\appendix

\section{Execution Structure Example}
\label{sec:appendix-structure}

Listing~\ref{lst:structure} shows the execution structure the Abstractor produced for the weak agent (\texttt{gpt-4.1-mini}) on SpreadsheetBench case 510-3 at iteration 0. Each entry of \texttt{nodes} is one procedure node $v_i$ of Eq.~\ref{eq:node-tuple}: \texttt{category\_tier1} is $c^1_i$ and \texttt{category\_tier2} is $c^2_i$, while \texttt{completes\_when} and \texttt{tool\_calls} are the supporting detail $d_i$, whose \texttt{code\_snippet} and \texttt{observation} are spliced in verbatim from the raw trajectory. The remaining fields are auxiliary: \texttt{id} and \texttt{label} let a gap report point at individual nodes (Listing~\ref{lst:gap}), \texttt{agent\_intent} and \texttt{purpose} record the agent's stated reason for the step and the call, \texttt{produces} and \texttt{consumes} give lightweight data flow, and \texttt{skill\_selection\_reason} carries the verbatim \texttt{reason} argument of an \texttt{activate\_skill} call, which is what makes wrong-skill routing detectable. Long snippets, observations, and absolute paths are abbreviated with \texttt{...} for space.

Node \texttt{node\_003} is the weak \textsc{validate} step that \S\ref{sec:localization} calls a \texttt{weak\_node}: the role is present, but its completion criterion stops at ``recalculated successfully without processing errors'' rather than reopening the workbook and checking the edited region.

\begin{lstlisting}[language=json,caption={Execution structure abstracted from a weak-agent trajectory (SpreadsheetBench case 510-3, iteration 0).},label={lst:structure}]
{
  "agent_id": "weak_iter0",
  "task": "Recreate the 'Data' sheet, preserve D:G formatting, hardcode Sum values, and save as output.xlsx",
  "total_steps": 5,
  "outcome": "success",
  "nodes": [
    {
      "id": "node_000", "label": "skill_activation",
      "category_tier1": "setup", "category_tier2": "skill_activation",
      "agent_intent": "Prepare the spreadsheet tooling needed to modify the Excel workbook.",
      "completes_when": "when the xlsx skill is activated and its spreadsheet handling instructions are available for use",
      "produces": ["xlsx_skill_context"], "consumes": [],
      "skill_selection_reason": "The task involves reading, modifying, and saving an Excel workbook while preserving formatting and hardcoding calculated values, which is best handled with the xlsx skill.",
      "tool_calls": [{
        "tool": "activate_skill",
        "purpose": "Activate the spreadsheet-editing skill required for workbook modification",
        "code_snippet": "{\"name\": \"xlsx\", \"reason\": \"The task involves reading, modifying, and saving an Excel workbook file ...\"}",
        "observation": "<activated_skill name=\"xlsx\"><instructions># Requirements for Outputs ...",
        "outcome": "success"}],
      "outcome": "success"
    },
    {
      "id": "node_001", "label": "task_requirement_review",
      "category_tier1": "setup", "category_tier2": "task_requirement_review",
      "agent_intent": "Understand the requested transformation and constraints before generating a script.",
      "completes_when": "when the sheet recreation requirements, formatting constraints, Sum calculation, and output filename are known",
      "produces": ["task_requirements"], "consumes": ["instruction_file"],
      "tool_calls": [{
        "tool": "read_file",
        "purpose": "Read the case instructions to capture the requested workbook edits and output requirements",
        "code_snippet": "{\"file_path\": \".../cases/510-3/case_input/INSTRUCTION.md\"}",
        "observation": "How can I calculate the total number of days from a date provided in column D of my 'Data' sheet ...",
        "outcome": "success"}],
      "outcome": "success"
    },
    {
      "id": "node_002", "label": "sheet_recreation_attempt",
      "category_tier1": "execute", "category_tier2": "sheet_recreation_attempt",
      "agent_intent": "Carry out the main workbook transformation by generating and running automation code.",
      "completes_when": "when a script has successfully recreated the Data sheet, preserved the required D:G formatting, hardcoded Sum values, and saved the modified workbook",
      "produces": ["transformation_script", "recreated_data_sheet", "hardcoded_sum_values", "output_workbook"],
      "consumes": ["task_requirements", "input_workbook"],
      "tool_calls": [{
        "tool": "write_file",
        "purpose": "Create a Python script to load the workbook, recreate the Data sheet, preserve formatting, calculate Sum values, and save",
        "code_snippet": "from openpyxl import load_workbook, Workbook\nfrom openpyxl.utils import get_column_letter\nimport openpyxl.styles as styles\n\ninput_path = ...",
        "observation": "Successfully wrote 3131 characters to .../weak_outputs/script.py",
        "outcome": "success"}, {
        "tool": "shell",
        "purpose": "Execute the generated Python script to produce the modified workbook",
        "code_snippet": "python3 .../weak_outputs/script.py",
        "observation": "exit_code: 0",
        "outcome": "success"}],
      "outcome": "success"
    },
    {
      "id": "node_003", "label": "workbook_recalculation_check",
      "category_tier1": "validate", "category_tier2": "workbook_recalculation_check",
      "agent_intent": "Validate that the produced workbook is usable after modification.",
      "completes_when": "when the generated workbook has been recalculated successfully without processing errors",
      "produces": ["recalculated_workbook_check"], "consumes": ["output_workbook"],
      "tool_calls": [{
        "tool": "shell",
        "purpose": "Recalculate the workbook to check whether the produced file opens cleanly and formulas or values evaluate without immediate errors",
        "code_snippet": "python3 .../working_skills/xlsx/scripts/recalc.py .../weak_outputs/output.xlsx",
        "observation": "stdout:\n{\"status\": \"success\", \"total_errors\": 0, \"total_formulas\": 0}\n\nexit_code: 0",
        "outcome": "success"}],
      "outcome": "success"
    },
    {
      "id": "node_004", "label": "output_workbook_save",
      "category_tier1": "output", "category_tier2": "output_workbook_save",
      "agent_intent": "Deliver the modified workbook in the requested output location.",
      "completes_when": "when output.xlsx exists in the output directory after the sheet recreation and Sum hardcoding steps",
      "produces": ["output.xlsx"], "consumes": ["output_workbook", "recalculated_workbook_check"],
      "tool_calls": [],
      "outcome": "success"
    }
  ]
}
\end{lstlisting}

\section{Gap Report Example}
\label{sec:appendix-gapreport}

Listing~\ref{lst:gap} shows one gap entry from a diagnoser gap report on SpreadsheetBench (\texttt{gpt-4.1-mini}, iteration 1; the full report for that iteration contains 15 such gaps). The prior iteration's patch had pushed the weak agent toward recalculating after an edit, but the weak agent still stops at a superficial check (recalc succeeding, or one unrepresentative cell) instead of reopening the workbook and spot-checking the target region; the diagnoser marks this gap \texttt{REGRESSION} (Table~\ref{tab:status}) across seven failing cases, and the \texttt{skill\_patch\_hint}'s PRIOR-PATCH CRITIQUE names four of them individually before prescribing a replacement verification protocol, following the TARGET/STYLE/WORDING/PRIOR-PATCH-CRITIQUE structure described in \S\ref{sec:diagnoser}.

\begin{lstlisting}[language=json,caption={One gap from a diagnoser gap report (\texttt{gap\_type=weak\_node}, \texttt{status=REGRESSION}).},label={lst:gap}]
{
  "gap_id": "gap_005",
  "gap_type": "weak_node",
  "sub_dimension": "no value-level spotcheck",
  "target_skill": "xlsx",
  "category_tier1": "validate",
  "category_tier2": "output_verification",
  "label": "recalc_without_targeted_readback_verification",
  "description": "The previous patch moved the agent toward recalculation, but current weak runs still often stop at recalc success, formula-only inspection, or a one-cell check instead of reopening the saved workbook and checking the actual visible results in the task-relevant region. This now appears in more failing cases than in iter 0, so the prior patch direction was right but measurably insufficient.",
  "weak_evidence": {
    "267-21": "python3 .../scripts/recalc.py .../weak_outputs/output.xlsx 30",
    "399-14": "print('Checking cells N12 and O12:', ws['N12'].value, ws['O12'].value)",
    "44628": "VALUES D4:G4: [20, 15, 35, 35]",
    "496-15": "ls -l .../weak_outputs/output.xlsx",
    "50631": "python3 .../scripts/recalc.py .../weak_outputs/output.xlsx",
    "50916": "print('Sample cell formulas and values:'); cells_to_check = ['C12','D12','E12','F12','G12','D13','E13','D14','E14']; [(c, ws_formulas[c].value, ws_values[c].value) for c in cells_to_check]",
    "57612": "FORMULA D4: =IF(AND(C4>=5,C4<=6),5,IF(AND(C4>=3,C4<=4),3.75,IF(AND(C4>=1,C4<=2),1,0)))"
  },
  "strong_evidence": {
    "267-21": "for r in ws.iter_rows(values_only=True): print(r)",
    "399-14": "for r in [12,15,16,18,21,22,7]: print(r, ws[f'N{r}'].value, ws[f'O{r}'].value, ws[f'P{r}'].value)",
    "44628": "print([ws[c].value for c in ['D3','E3','F3','G3']])",
    "496-15": "for row in range(1, ws.max_row + 1): print(row, d, e)",
    "50631": "for coord in ['B36','B37','B38','D7','D38','F7','F38','H7','H30','H36','H37','H38']: print('value', coord, ws2[coord].value)",
    "50916": "for r in [12,13,14,17,18,19,20,21]: print(r, [ws.cell(r,c).value for c in range(1,9)])",
    "57612": "for r in range(4,16): print(r, wsv[f'C{r}'].value, wsv[f'D{r}'].value)"
  },
  "strong_node_ref": {"267-21": "node_007", "399-14": "node_004", "44628": "node_004",
    "496-15": "node_005", "50631": "node_004", "50916": "node_004", "57612": "node_004"},
  "weak_node_ref": {"267-21": "node_009", "399-14": "node_005", "44628": "node_006",
    "496-15": "node_003", "50631": "node_006", "50916": "node_005", "57612": "node_004"},
  "strong_completes_when": "when the saved workbook has been reopened in both formula and data-only views and the task-relevant edited region has been explicitly spot-checked",
  "weak_completes_when": "when recalc succeeds, the file exists, or a tiny unrepresentative slice looks plausible, without verifying the actual target range or final visible values",
  "missing_produces": ["verification_report", "spotcheck_results"],
  "skill_patch_hint": "TARGET: patch xlsx/SKILL.md in the Recalculating formulas section and the Common Workflow after recalc. STYLE: MANDATE, and because this gap is REGRESSION, explicitly replace the prior weak mechanism rather than merely intensifying it. WORDING: add a post-recalc hard-stop checklist saying the agent must reopen the canonical output.xlsx twice: once with data_only=False to verify formulas or structural edits landed in the intended addresses, and once with data_only=True to verify visible values in the actual destination region. Require four checks: the named target cells from the prompt, the first edited row, one middle or edge row, and the final edited or summary row. Also require one unchanged control row or cell to confirm no unintended spillover. Add a hard-stop sentence: 'A zero-error recalc report, save message, or file-exists check is not completion. If readback shows wrong placement, blank results, untouched ranges, wrong totals, or the wrong sheet layout, repair and re-verify.' PRIOR-PATCH CRITIQUE: the iter-0 patch successfully made recalc common, but current structures show the agent satisfies the letter while skipping the real semantic check. 496-15 stops at ls -l. 50631 ends after recalc and an H38 patch without checking the full date extent. 50916 executes a list comprehension that prints nothing, then claims success. 44628 reads back the wrong cells and accepts D4:G4 instead of D3:G3. The new patch must therefore switch mechanisms from generic recalc guidance to a mandatory readback protocol with explicit target-region coverage.",
  "failed_case_ids": ["267-21", "399-14", "44628", "496-15", "50631", "50916", "57612"],
  "pass_case_ids": ["84-40"],
  "status": "REGRESSION"
}
\end{lstlisting}

\section{Validator Report Example}
\label{sec:appendix-validator}

Listing~\ref{lst:validator} shows a validator report for a SpreadsheetBench case (48608, iteration 0). The validator compares the weak agent's \texttt{output.xlsx} against the strong agent's own output, not the gold answer (\S\ref{sec:diagnoser}). Its top-level \texttt{verdict} is PASS only if every expected file matches exactly; PARTIAL if a file is present but imperfect (as here) or one expected file is entirely missing; FAIL if most expected files are missing or severely wrong. The \texttt{scores} block gives \texttt{file\_presence\_ratio} (share of expected files present in any form), \texttt{content\_match\_ratio} (share of present files that are an exact match), and their weighted blend \texttt{overall\_quality\_score} $=0.4\times\text{presence}+0.6\times\text{content}$. Within a file's \texttt{detail}, \texttt{row\_match\_ratio} counts whole strong rows found anywhere in the weak output, \texttt{cell\_match\_ratio} is the finer-grained cell-by-cell match, and \texttt{per\_column\_match\_ratio} breaks the latter down by column. Here, the file is present with matching sheet structure, but a 0.0 match ratio on the \textsc{Passes} column, yielding \texttt{verdict=PARTIAL}.

\begin{lstlisting}[language=json,caption={Validator report for case 48608, iteration 0 (\texttt{verdict=PARTIAL}).},label={lst:validator}]
{
  "strong_agent": "strong",
  "weak_agent": "weak_iter0",
  "verdict": "PARTIAL",
  "result_quality": {
    "files": {
      "present": [
        {
          "filename": "output.xlsx",
          "matched_weak_filename": "output.xlsx",
          "content_match": false,
          "comparator": "xlsx",
          "detail": {
            "sheets_match": true,
            "strong_sheets": ["Sheet1"],
            "weak_sheets": ["Sheet1"],
            "cell_match_ratio": 0.9091,
            "per_sheet": {
              "Sheet1": {
                "present": true,
                "row_match_ratio": 0.125,
                "cell_match_ratio": 0.9091,
                "strong_rows": 8,
                "weak_rows": 17,
                "missing_rows": 7,
                "extra_rows": 16,
                "per_column_match_ratio": {
                  "Name": 1.0,
                  "Session": 1.0,
                  "Result": 1.0,
                  "": 1.0,
                  "Passes (Hidden)": 1.0,
                  "Passes": 0.0
                }
              }
            }
          }
        }
      ],
      "missing_from_weak": [],
      "extra_in_weak": []
    },
    "scores": {
      "file_presence_ratio": 1.0,
      "content_match_ratio": 0.0,
      "overall_quality_score": 0.4
    }
  },
  "analysis": "The weak agent produced the required workbook file, and the workbook structure matches the reference sheet name. However, the sheet contents do not fully match the reference. The overall cell match is high, but the key 'Passes' column does not match the reference, and the row alignment is poor, with many extra and missing rows. This indicates a partially correct spreadsheet output rather than an exact reproduction.",
  "verdict_reasoning": "There is one expected file and it is present, so file presence is complete. The content is not an exact match due to differences in the workbook rows and the 'Passes' column, so content_match_ratio is 0. Because the output is present but not correct, the verdict is PARTIAL."
}
\end{lstlisting}

\section{Adapted Skill Example}
\label{sec:appendix-skilldiff}

Listing~\ref{lst:skilldiff} shows the full diff between the original xlsx skill $\mathcal{S}_0$ and the skill adapted for \texttt{gpt-5.4-mini} on SpreadsheetBench: four hunks, 63 lines removed and 79 added. The edits fall into three kinds: unconditional rules become conditional on what the prompt asks, vague instructions become explicit scoping steps that must precede an edit, and generic verification advice becomes a mandatory readback gate on the saved workbook together with rules for what to do when it fails.

\begin{lstlisting}[language=skilldiff,caption={\texttt{skill\_changes\_xlsx.diff} for \texttt{gpt-5.4-mini} on SpreadsheetBench. Header paths are shortened and long lines are cut at a sentence boundary, marked \texttt{[...]}.},label={lst:skilldiff}]
--- skill_baseline_xlsx/SKILL.md
+++ skill_final_xlsx/SKILL.md
@@ -15,8 +15,9 @@
 - Every Excel model MUST be delivered with ZERO formula errors (#REF!, #DIV/0!, #VALUE!, #N/A, #NAME?)
 
 ### Preserve Existing Templates (when updating templates)
-- Study and EXACTLY match existing format, style, and conventions when modifying files
-- Never impose standardized formatting on files with established patterns
+- Inspect and EXACTLY match the workbook's structure, styles, merges, dimensions, and conventions before editing
+- If the prompt says recreate, rebuild, or preserve a sheet exactly, capture the template first and treat sheet copy/recreation as a required step before value fixes
+- Never treat in-place edits as equivalent to recreation when template fidelity, split fields, or preserved layout matter
 - Existing template conventions ALWAYS override these guidelines
 
 ## Financial models
@@ -96,57 +97,26 @@
 
 ## Excel File Workflows
 
-## CRITICAL: Use Formulas, Not Hardcoded Values
+## CRITICAL: Choose formulas vs hardcodes explicitly
 
-**Always use Excel formulas instead of calculating values in Python and hardcoding them.** This ensures the spreadsheet remains dynamic and updateable.
-
-### WRONG - Hardcoding Calculated Values
-```python
-# Bad: Calculating in Python and hardcoding result
-total = df['Sales'].sum()
-sheet['B10'] = total  # Hardcodes 5000
-
-# Bad: Computing growth rate in Python
-growth = (df.iloc[-1]['Revenue'] - df.iloc[0]['Revenue']) / df.iloc[0]['Revenue']
-sheet['C5'] = growth  # Hardcodes 0.15
-
-# Bad: Python calculation for average
-avg = sum(values) / len(values)
-sheet['D20'] = avg  # Hardcodes 42.5
-```
-
-### CORRECT - Using Excel Formulas
-```python
-# Good: Let Excel calculate the sum
-sheet['B10'] = '=SUM(B2:B9)'
-
-# Good: Growth rate as Excel formula
-sheet['C5'] = '=(C4-C2)/C2'
-
-# Good: Average using Excel function
-sheet['D20'] = '=AVERAGE(D2:D19)'
-```
-
-This applies to ALL calculations - totals, percentages, ratios, differences, etc. The spreadsheet should be able to recalculate when source data changes.
+- If the task asks to build, fix, copy, extend, summarize, or preserve spreadsheet behavior -- or the workbook should update when source cells change -- write Excel formulas into the workbook, not Python-computed results
+- If the prompt explicitly asks for static final values, hardcoded outputs are allowed because the user requested a non-dynamic result
+- Do not choose Python-computed cell values merely because they are easier when workbook logic is part of the deliverable
+- Dynamic example: write a lookup, SUMIF, or fill-down formula into the target cells
+- Static exception example: the prompt explicitly asks for final pasted values with no formulas
+- When the dynamic branch applies, reopen the saved workbook with normal loading and confirm the target cells contain formulas, not just displayed numbers
 
 ## Common Workflow
 1. **Choose tool**: pandas for data, openpyxl for formulas/formatting
 2. **Create/Load**: Create new workbook or load existing file
-3. **Modify**: Add/edit data, formulas, and formatting
-4. **Save**: Write to file
-5. **Recalculate formulas (MANDATORY IF USING FORMULAS)**: Use the scripts/recalc.py script
+3. **Audit target before editing**: identify the editable deliverable sheet, any source-data sheets, and any example/reference/oracle sheets before changing cells. Do not overwrite sheets named like Desired Results, Example, Template, or Reference unless the prompt explicitly names them as the sheet to modify. Then identify the exact destination range, condition or lookup columns, first and last affected row, and whether the task is whole-column, bounded-range, table, merged-region, summary-section, or sheet-recreation work. If the task deletes, filters, copies, or keeps rows based on only some columns inside a wider visible band, write down the exact named condition columns first and treat adjacent numbering, helper, or metadata columns as out of scope unless the prompt names them explicitly. If the destination subset is smaller than the source table, derive destination start/end rows separately from source lookup bounds. For existing summary blocks, identify the label column and the result column before writing anything; prove the mapping on the first label row and on one later label row so you know the real destination cells. Check at least one sample near the start and one near the end of the intended span. Do not infer bounds from the prompt alone or from a tiny preview. If the prompt says recreate or preserve a sheet exactly, inspect styles, merged cells, and layout anchors before touching data
+4. **Modify the workbook state**: apply the requested spreadsheet change itself; do not substitute code text, macro text, instructions, comments, or note cells as the main solution. For formula repairs, copied patterns, schedule grids, or repeated rules, state the intended destination start row, end row, start column, and end column before writing, then apply the change across that exact region. If the prompt names one cell as an example but describes a whole column, repeated rule, copied formula pattern, or schedule grid, derive the workbook-defined destination span from the target area itself and fill every target row in that span, e.g. `for row in range(start_row, end_row+1): if ws[f'{driver_col}{row}'].value is not None: ws[f'{target_col}{row}'] = make_formula(row)`. For report strips, pass lists, or other bounded destinations, destination bounds are separate from source lookup bounds. If a row-selection rule depends on a named subset of columns, build the predicate from that exact subset rather than the whole inspected slice: bad = checking every visible column because it is nearby; good = checking only the explicitly named subset, then applying deletion/filter/copy logic from that predicate. Read back one kept-row boundary and one deleted-row boundary to prove the subset-column rule, not just the final row count. If merged cells block direct fill, build the full editable non-merged target set and fill every editable target cell across the whole intended grid; merged cells are not permission to retreat to a tiny safe subset. For workbook summary sections, preserve the existing label/result layout: if rows already contain labels like `Sum A`, `Sum B`, or similar, write formulas into the established result cells adjacent to those labels rather than into a nearby blank column. Verify the same label-to-result offset on more than one summary row before broad fill. For schedule/calendar formulas, classify each reference before fill as fixed-row, fixed-column, or relative. Keep the day/cycle key on the current row in the day column, let the return range move with the destination period column, and build one anchor formula that survives both horizontal and vertical translation, e.g. `=IFERROR(INDEX(C$2:C$8,MATCH($B12,$B$2:$B$8,0)),"")`; then prove it on the anchor cell, one horizontally shifted neighbor, and one next-row cell before broad fill. Do not translate a legacy nested-IF or copied formula whose anchor already points at the wrong row, header, day field, or prior output cell, and do not build chain formulas that match against earlier filled schedule cells. For mixed text-and-number summary cells like category codes with numeric suffixes, build the formula from the actual data-cell range itself, not from summary labels or an oversized sheet rectangle: compare the category token to `LEFT(source_range,1)`, parse the numeric suffix from that same `source_range`, and sum with an error-tolerant conversion so blanks and nonmatches contribute zero. Do not compare whole cells in a broad block to a bare category letter, and do not parse suffixes from label rows or blank-leading regions. If readback or validation shows the visible divergence is in a repeated date-series column, rebuild that exact date family first from an explicit row offset from the first target row, e.g. `=IF(start_date+offset<=end_date,start_date+offset,"")`, rather than repairing a nearby formula family that merely coexists in the workbook. After rebuilding the date family, verify the first date, the last intended date, and the first blank cell after the series. One-cell or sample-cell patches are not sufficient unless the prompt explicitly names one cell as the entire target
+5. **Save**: Write to file
+6. **Recalculate formulas (MANDATORY IF USING FORMULAS)**: Use the scripts/recalc.py script
    ```bash
    python scripts/recalc.py output.xlsx
    ```
-6. **Verify and fix any errors**: 
-   - The script returns JSON with error details
-   - If `status` is `errors_found`, check `error_summary` for specific error types and locations
-   - Fix the identified errors and recalculate again
-   - Common errors to fix:
-     - `#REF!`: Invalid cell references
-     - `#DIV/0!`: Division by zero
-     - `#VALUE!`: Wrong data type in formula
-     - `#NAME?`: Unrecognized formula name
+7. **Read back the final file (MANDATORY ACCEPTANCE GATE)**: after every edit, reopen the saved workbook and verify the exact target cells against the actual task invariant; do not use readback as logging only. If formulas were inserted or changed, run recalc first, then use `data_only=True` to check final values and normal loading when formula text or formatting also matters. For small edited regions, check every target output cell. For large edited regions, check the first edited cell, the last edited cell, one boundary case, and one logic-proving case that can expose the likely failure mode. Mark each check `PASS` or `FAIL` against the business rule, not merely against "cell has a value," "format looks right," or "file opens." Treat readback as a go/no-go gate: if a class cell reads `False` or blank where a class label should appear, a repaired summary reads all zeros despite visible matching source data, a latest-four result disagrees with a manual row check, a date-series checkpoint still contradicts the requested progression, the first untouched row now contains the filled formula, or any other target invariant fails, the workbook is not deliverable. Return to editing, repair the failing logic, and repeat save -> recalc -> readback. The agent MUST NOT deliver the workbook until every readback check passes
 
 ### Creating new Excel files
 
@@ -204,6 +174,22 @@
 wb.save('modified.xlsx')
 ```
 
+Before modifying an existing workbook:
+- Identify sheet roles first: which sheet is the editable deliverable target, which sheets are only source data, and which sheets are examples, reference tabs, templates, or desired-results/oracle tabs
+- Use example/reference/desire-results sheets to infer the required transformation or compare outcomes, but do not overwrite them unless the prompt explicitly tells you to edit that sheet; bad: copying the desired ordering into the reference sheet, good: reordering the working sheet so it matches the reference arrangement
+- Derive the full affected region from the workbook itself: source/destination sheet, start row, end row, start column, end column, stopping rule, and any condition columns or table bounds
+- If row deletion, filtering, or copying depends on only some columns inside a wider band, identify the exact named subset columns before editing and keep nearby numbering/helper columns out of the predicate unless the prompt explicitly includes them
+- If the task says recreate a sheet, rebuild a template, or preserve a tab exactly, duplicate or rebuild the target structure first, preserve row/column layout plus style anchors and merges, then populate repaired values or formulas
+- If the prompt names one cell but describes a repeated rule, copied formula pattern, or whole target column, derive the workbook-defined first/last target rows plus start/end columns from the destination area and fill every row in that span rather than only the named example; state `start_row`, `end_row`, `start_col`, and `end_col` before writing. Reusable pattern: `for row in range(start_row, end_row+1): if ws[f'{driver_col}{row}'].value is not None: ws[f'{target_col}{row}'] = template_for(row)`
+- If merged cells block direct fill, enumerate the full editable non-merged target set across the intended grid and fill every editable target cell; encountering merged cells is not permission to abandon the rest of the grid
+- For existing summary sections, identify the label column and the result column from the sheet structure before writing formulas. Prove the destination mapping on the first label row and one later label row, then write into those established result cells rather than a nearby blank column
+- For schedule/calendar formulas, classify references before filling: fixed-column, fixed-row, or relative. Keep the match key on the current row in the day column, let the return range move with the destination period column, and build one anchor formula that translates correctly in both directions, e.g. `=IFERROR(INDEX(C$2:C$8,MATCH($B12,$B$2:$B$8,0)),"")`; then test the anchor cell, one horizontally shifted neighbor, and one next-row cell before broad fill
+- Do not translate a legacy nested-IF or copied formula whose anchor already points at the wrong field (for example the wrong row, header, date column, day field, or prior output cell); translation spreads wrong logic faithfully
+- If writing category totals from mixed cells like A2, B1, or F3, build the formula from the actual data-cell range itself, not from summary labels or the whole sheet rectangle. Compare `LEFT(source_range,1)` to the category token, parse the numeric suffix from that same `source_range`, and sum with an error-tolerant conversion so blanks and nonmatches contribute zero; generic pattern: match the prefix from `source_range`, extract the suffix from `source_range`, and aggregate only those parsed matches
+- Do not compare whole cells in a broad block to a bare category letter, and do not parse suffixes from label rows or blank-leading regions; those anti-patterns collapse the logic to mismatches and zero totals
+- Zero totals across every summary row are a red flag when the visible source table contains matching mixed letter-number entries
+- If only a subset of rows belongs in the output, derive destination start/end rows separately and bound the fill to that subset rather than extending to the whole sheet by default
+
 ## Recalculating formulas
 
 Excel files created or modified by openpyxl contain formulas as strings but not calculated values. Use the provided `scripts/recalc.py` script to recalculate formulas:
@@ -224,27 +210,57 @@
 - Returns JSON with detailed error locations and counts
 - Works on both Linux and macOS
 
+Recovery rules after recalc:
+- If `status` is `errors_found`, inspect `error_summary`, fix the specific cells, and rerun recalc immediately
+- Do not perform multiple near-identical formula rewrites that keep the same failing coercion or parsing pattern
+- After the second recalc failure on the same target cells, stop whole-range rewrites. Make one real mechanism switch on one pilot target cell, prove it on the saved workbook, and only then fill the region
+- Do not cycle through TAKE/FILTER, LARGE/IF, SUMPRODUCT-with-LARGE, AGGREGATE, or similar array-heavy variants that all chase the same underlying idea without first proving one engine-compatible mechanism on the pilot cell. Superficial syntax swaps do not count as a new mechanism
+- Change mechanism based on the error class; for mixed text/number parsing across ranges, use an error-tolerant formula pattern so blanks, unmatched cells, or nonnumeric entries contribute zero instead of propagating errors
+- For row-wise "latest N nonblank values" tasks, do not use `COUNT(range)` alone to choose a contiguous tail slice unless the workbook guarantees nonblank entries are left-packed. When blanks may appear between valid entries, derive the cutoff position of the Nth-latest populated cell first from a nonblank-position mask such as `LARGE((range<>"")*COLUMN(range),MIN(N,COUNT(range)))`, then aggregate only cells at or after that cutoff with a scalar pattern such as `SUMPRODUCT((range<>"")*(COLUMN(range)>=cutoff)*range)/MIN(N,COUNT(range))`
+- For latest-N recovery, write the candidate formula to one representative row, run recalc, reopen with `data_only=True`, compare the numeric result to a manual latest-N check for that row, and only then fill the remaining rows
+- Do not attempt a third retry unless the mechanism has actually changed and the pilot-cell proof succeeded
+- After recalc succeeds, still reopen the saved workbook and read back one or two target cells to confirm no formula errors and plausible values
+
 ## Formula Verification Checklist
 
 Quick checks to ensure formulas work correctly:
 
-### Essential Verification
-- [ ] **Test 2-3 sample references**: Verify they pull correct values before building full model
-- [ ] **Column mapping**: Confirm Excel columns match (e.g., column 64 = BL, not BK)
-- [ ] **Row offset**: Remember Excel rows are 1-indexed (DataFrame row 5 = Excel row 6)
-
-### Common Pitfalls
-- [ ] **NaN handling**: Check for null values with `pd.notna()`
-- [ ] **Far-right columns**: FY data often in columns 50+ 
-- [ ] **Multiple matches**: Search all occurrences, not just first
-- [ ] **Division by zero**: Check denominators before using `/` in formulas (#DIV/0!)
-- [ ] **Wrong references**: Verify all cell references point to intended cells (#REF!)
-- [ ] **Cross-sheet references**: Use correct format (Sheet1!A1) for linking sheets
-
-### Formula Testing Strategy
-- [ ] **Start small**: Test formulas on 2-3 cells before applying broadly
-- [ ] **Verify dependencies**: Check all cells referenced in formulas exist
-- [ ] **Test edge cases**: Include zero, negative, and very large values
+### Scope before writing
+- [ ] Derive the exact target span from the workbook: sheet, start/end row, start/end column, stopping rule, and any condition/lookup columns
+- [ ] If the destination subset is smaller than the source table, derive destination start/end rows separately and state them before filling
+- [ ] Check at least one sample near the start and one near the end of the intended span
+- [ ] Confirm column mapping and row offsets before filling formulas
+- [ ] If preserving or recreating a template, inspect styles, merges, and layout anchors before editing
+
+### Building and filling formulas
+- [ ] Test 2-3 sample references before broad fill
+- [ ] Verify dependencies and cross-sheet references exist
+- [ ] State the destination `start_row`, `end_row`, `start_col`, and `end_col` before writing any repeated-rule fix
+- [ ] If the prompt names an exemplar cell but describes a repeated rule, fill the full workbook-defined region rather than just the sample cell
+- [ ] If a row condition uses only some columns in a wider visible band, define and use that exact named subset rather than the whole slice
+- [ ] If merged cells exist, enumerate all editable non-merged targets in the intended grid and fill all of them
+- [ ] For existing summary rows, identify the label column and the result column, prove the offset on the first label row and one later label row, and write only into those established target cells
+- [ ] For copied calendar/schedule formulas, decide fixed-column vs fixed-row vs relative references before fill; keep the match key on the current row day column, let the return range move with the destination period column, and test one anchor cell, one horizontally shifted neighbor, and one next-row cell
+- [ ] Do not copy a schedule formula whose anchor already points at the wrong field or a prior output cell
+- [ ] For mixed text+number category summaries, compare the category token to the prefix from the real data-cell range and parse the numeric suffix from that same range; do not aggregate from label rows or oversized sheet rectangles
+- [ ] For latest-N-nonblank logic, reject a `COUNT(range)`-based contiguous tail unless nonblank cells are guaranteed to be left-packed; prove the formula selects positions of the last N populated cells
+- [ ] Before broad fill after a formula-family repair, prove one pilot row against a manual check on the recalculated workbook
+- [ ] Bound the fill to the true target subset; do not extend to unrelated rows by default
+- [ ] Use error-tolerant patterns when blanks or nonnumeric text may appear
+
+### Readback after save
+- [ ] Run `scripts/recalc.py` after formula changes
+- [ ] Reopen with `data_only=True` for final values and normal loading when formula text or formatting must also be checked
+- [ ] Compare observed exact target cells to the task-specific acceptance rule, not just to "file opens," "format looks right," or "cell has something"
+- [ ] Small target: check every output cell. Large target: check the first edited cell, the last edited cell, one boundary case, and one logic-proving case
+- [ ] Mark every readback check `PASS` or `FAIL` explicitly
+- [ ] Treat `False` or blank class cells where labels should appear, all-zero summaries despite visible matching source data, latest-four results that fail a manual row check, date progressions that still diverge, content mismatches hidden by correct formatting, `#VALUE!`, `None`, wrong booleans, numerically wrong invariants, or a changed first-untouched row as failure signals that require more editing before delivery
+- [ ] For summary blocks, read back the real destination cells, not a nearby column that merely contains plausible computed values
+- [ ] If repeated date-series columns still diverge, repair that exact date family first and re-check the first date, last intended date, and first blank after the series
+- [ ] For row-deletion/filter tasks based on subset columns, read back one kept-row boundary and one deleted-row boundary that prove the predicate used the right columns
+- [ ] For pilot-cell recovery after repeated recalc failures, prove one target cell calculates correctly before filling the rest of the region
+- [ ] For dynamic-workbook tasks, confirm the target cells still contain formulas, not hardcoded numbers
+- [ ] For recreated sheets, verify both preserved structure/formatting and repaired content fields
 
 ### Interpreting scripts/recalc.py Output
 The script returns JSON with error details:
\end{lstlisting}

% \newpage
% \input{checklist.tex}

\end{document}